\documentclass[conference]{IEEEtran}
\IEEEoverridecommandlockouts
\usepackage{cite}
\usepackage{amsmath,amssymb,amsfonts}
\usepackage{algorithmic}
\usepackage{textcomp}
\usepackage{xcolor}
\usepackage{todonotes}
\usepackage{float}
\usepackage{url}
\usepackage{filecontents}
\def\BibTeX{{\rm B\kern-.05em{\sc i\kern-.025em b}\kern-.08em
    T\kern-.1667em\lower.7ex\hbox{E}\kern-.125emX}}

\usepackage{graphicx}

\usepackage[english]{babel}

\begin{document}

\title{Synthetic Data Augmentation for Table Detection:\\
Re-evaluating TableNet's Performance with\\
Automatically Generated Document Images}

\author{\IEEEauthorblockN{Krishna Sahukara, Zineddine Bettouche, Andreas Fischer}
    \IEEEauthorblockA{Deggendorf Institute of Technology \\
        Dieter-Görlitz-Platz 1, 94469 Deggendorf \\
        krishna.sahukara@stud.th-deg.de, \{zineddine.bettouche, andreas.fischer\}@th-deg.de}
}

\maketitle

\begin{abstract}
    Document pages captured by smartphones or scanners often contain tables, yet manual extraction is slow and error-prone. We introduce an automated \emph{LaTeX}-based pipeline that synthesises realistic two-column pages with visually diverse table layouts and aligned ground-truth masks. The generated corpus augments the real-world Marmot benchmark and enables a systematic resolution study of TableNet. Training TableNet on our synthetic data achieves a pixel-wise XOR error of 4.04~\% on our synthetic test set with a 256\,\texttimes{}256 input resolution, and 4.33~\% with 1024\,\texttimes{}1024. The best performance on the Marmot benchmark is 9.18~\% (at 256\,\texttimes{}256), while cutting manual annotation effort through automation.
\end{abstract}

\begin{IEEEkeywords}
    document image analysis, deep learning, synthetic data generation, table structure recognition, semantic segmentation, TableNet, performance evaluation
\end{IEEEkeywords}

\section{Introduction}

Smartphones and scanners are used to capture images of documents containing tables with vital information. Manual extraction is time-consuming. However, a shortage of annotated datasets has made developing effective deep learning models for table detection challenging.

Data augmentation has been recognized as a critical technique in deep learning research, with surveys highlighting its effectiveness in addressing training data limitations and enhancing model generalization capabilities~\cite{survey1, survey2}. While these studies explore numerous augmentation strategies across various domains, tabular text data in document images presents unique challenges not fully addressed in existing literature. Our synthetic data generation approach draws inspiration from generative modeling principles, similar to those in our work on process mining~\cite{anjali}, but is tailored for the structural and visual complexities of tabular document images.

To address this challenge, we propose creating a specialized dataset for table detection. Our method automates the annotation process, reducing manual effort and ensuring a diverse dataset for robust model training. By employing deep learning, we aim to develop a system capable of autonomously identifying and extracting text from tabular regions in document images.

We explore an end-to-end deep learning model that integrates table detection and structure identification, streamlining the process and improving efficiency. This model treats table detection and structure identification as interconnected tasks, enhancing accuracy while reducing computational demands, making it suitable for real-world applications with resource constraints.

Our approach improves table detection by combining a newly generated dataset with a resolution-aware TableNet baseline.  The resulting models facilitate downstream tasks such as large-scale document processing and information retrieval.

\paragraph*{Contributions}The main contributions of this paper are:
\begin{itemize}
    \item an automated LaTeX pipeline that renders page-level synthetic documents with diverse table styles and aligned ground-truth masks;
    \item a public corpus and code base that enable reproducible training and evaluation;
    \item an extensive comparison of TableNet at two input resolutions (256\texttimes256 vs.~1024\texttimes1024) to analyze performance trade-offs;
\end{itemize}

As for the structure of this paper, Section~\ref{background} introduces the technologies used, Section~\ref{relatedWork} surveys prior work, Section~\ref{methodology} details the dataset and evaluation protocol, Section~\ref{sec:implementation} describes the implementation, Section~\ref{sec:results} presents the results and their discussion, and Section~\ref{sec:conclusion} concludes the paper and outlines future directions.

\section{Background}\label{background}
This section presents the background of the used networks, VGG-19 and TableNet.

\subsection{VGG-19}
VGG-19~\cite{simonyan2014very} is a 19-layer network that stacks 3\,\texttimes{}3 convolutions and 2\,\texttimes{}2 max-pooling operations. Despite its age, its simple design and publicly available ImageNet weights make VGG-19 a convenient backbone for feature extraction; we therefore adopt it as the shared encoder in our TableNet baselines.

\subsection{Image Segmentation}
Semantic segmentation~\cite{guo2018review} assigns a semantic label to every pixel instead of producing coarse bounding boxes. State-of-the-art methods are fully convolutional and benefit from encoder weights pre-trained on large classification corpora, enabling accurate dense predictions even when task-specific training data are limited.

\subsection{TableNet}
Earlier methods in deep learning treated table and column detection as separate issues. However, searching for columns independently often led to false positives due to their vertical alignment, complicating accurate detection. Leveraging knowledge of the tabular region improved column detection by using convolution filters to detect both tables and columns. 

TableNet~\cite{paliwal2019tablenet}, utilizing an encoder-decoder model for semantic segmentation, capitalized on this by employing the same encoder for both tasks and separate decoders for each. The architecture incorporates pre-trained VGG-19 features, shared encoder layers, and specific decoder branches for table and column segmentation. Each branch uses convolutional layers and transposed layers for feature map dimension reduction and pixel class prediction.

\section{Related Work}\label{relatedWork}

Various studies and surveys in table comprehension~\cite{couasnon2014recognition, zanibbi2004survey, silva2006design, khusro2015methods} have explored table identification and data extraction, often reporting results separately~\cite{embley2006table}. Prior to deep learning, table identification relied on heuristics or metadata. Early efforts, such as TINTIN~\cite{pyreddy1997tintin} and the work by Cesarini et al.~\cite{cesarini2002trainable}, involved structural information and machine learning techniques for table identification, using MXY trees and hierarchical representations with Tabfinder.

Other approaches, like that of T. Kasar et al.~\cite{kasar2013learning}, focused on detecting intersecting lines and employing SVM classifiers, but were limited by visible guidelines. Probabilistic models, such as those by Silva et al.~\cite{silva2009learning}, emphasized combining multiple approaches using joint probability distribution and hidden row states.

While table detection received significant attention, table structure identification garnered less focus. Notable works include the T-RECS method by Kieninger and Dengel~\cite{kieninger1999t}, Wang et al.'s seven-step process for table structure understanding~\cite{wang2004table}, and Shigarov et al.'s system that incorporated extensive configuration options and PDF metadata~\cite{shigarov2016configurable}.

Recent advancements have integrated deep learning into table detection and structure recognition, as seen in methods by Hao et al.~\cite{hao2016table} and Tran et al.~\cite{tran2015table}, both achieving competitive performance. DeepDeSRT~\cite{schreiber2017deepdesrt, kavasidis2018saliency} and works by Aswin et al.~\cite{tengli2004learning} and Singh et al.~\cite{singh2018multidomain} have leveraged deep learning and object detection techniques to improve table detection and structure recognition, showing advancements in performance and data extraction from different document formats.

Recent large-scale \,\emph{synthetic} corpora such as SynthTabNet~\cite{synctabnet2021}, PubTables-1M~\cite{pubtables1m2022}, and DocBank~\cite{docbank2021} have accelerated research on table understanding. A common trait of these resources is that released images typically contain an \,\textit{isolated} table crop rather than the entire document page. While this facilitates fast training for cell-structure recognition, it provides limited context for joint layout analysis. Our automatically rendered dataset, in contrast, embeds tables within two-column scientific pages so models must reason about neighbouring text, captions, and figure regions in addition to the table area itself.

Alongside new datasets, recent architectures have improved performance. Transformer-based detectors such as TableTransformer~\cite{tabletransformer2021} and encoder--decoder approaches like TableFormer~\cite{tableformer2021} achieve state-of-the-art results on PubTabNet, whereas two-stage detectors such as Cascade Mask R-CNN~\cite{cai2019cascade} remain strong contenders in dense-layout benchmarks. Rather than benchmarking every available method, we deliberately focus on a representative encoder--decoder baseline (TableNet) to isolate the effect of data augmentation and input resolution; integrating complementary detectors is left for subsequent studies.

\section{Methodology}\label{methodology}

This section presents the dataset, our generation approach, and the evaluation methods.

\subsection{Used Datasets: Marmot}
The Marmot dataset~\cite{marmotdataset} is a large-scale collection of document images commonly used for table detection. It serves as a tool for assessing table detection algorithms in real-world scenarios. The dataset maintains a near 1:1 positive-to-negative ratio, covering various document types such as research papers and forms. Its authenticity underscores its relevance for training models on real-world challenges.

The Marmot dataset suffers from labour-intensive manual annotations that are time-consuming and costly.

\subsection{The Proposed Generation Approach: Novel Training Data}
We propose an approach to automatically generate a wide range of training data.

Our evaluation comprises both synthetic-to-synthetic and synthetic-to-real testing scenarios: models are validated on the held-out portion of our generated corpus and, independently, on the external Marmot benchmark. This dual protocol enables us to disentangle domain-specific effects from purely architectural factors.

The proposed approach takes as input 4 parameters: number of rows, number of columns, datatypes, and table style. These are saved as ground-truth for the evaluation. Employing LaTeX and Lorem Ipsum in a Python script, the output is an image used in training table detection models.

\begin{figure}
    \centering
	  \includegraphics[width=.8\linewidth]{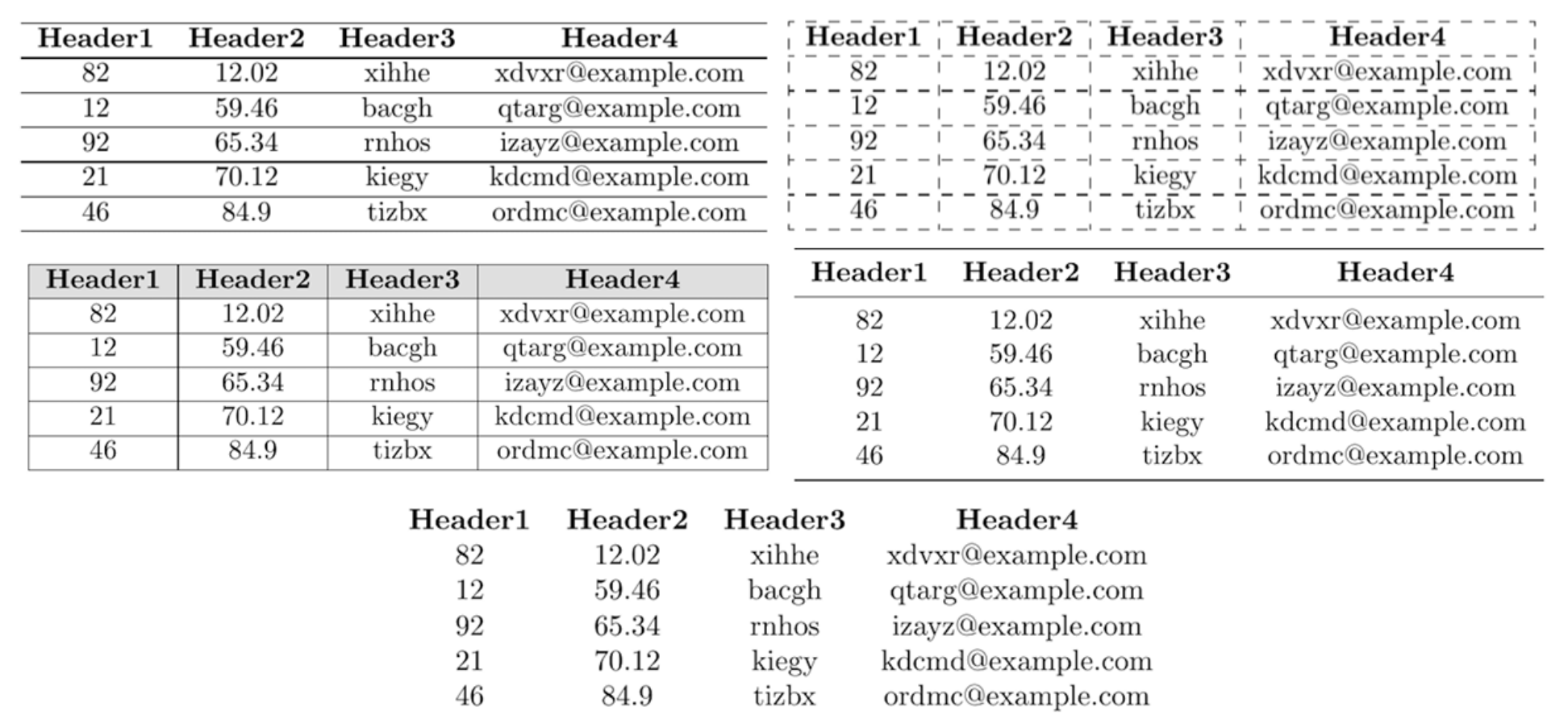}
    \caption{Styles of Generated Tables: dashed lines, without vertical borders, colored headers, devoid of top/mid/bottom rules, and booktabs.}
    \label{fig:table-styles}
\end{figure}

\begin{figure}
    \centering
	  \includegraphics[width=.8\linewidth]{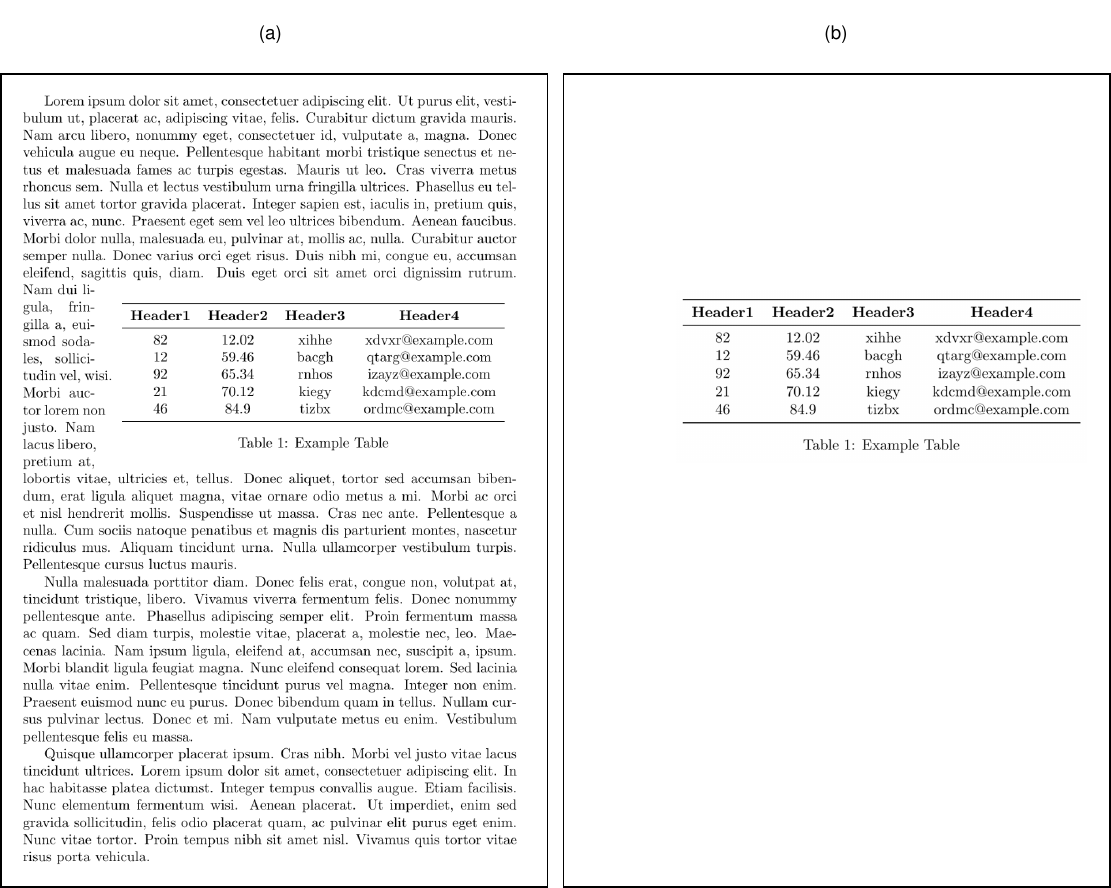}
    \caption{Types of Generated Image: (a) with text; (b) without text.}
    \label{fig:generated-image}
\end{figure}

The generated tables fall into five different styles (Fig.~\ref{fig:table-styles}). Each autogenerated table is populated with statistically valid dummy data matching column datatypes: numerical columns contain Gaussian-distributed values ($\mu=0$, $\sigma=1$), text columns use Markov chain-generated Lorem Ipsum, and date fields follow a uniform distribution across 2000–2023. The generated images are of two types: a table surrounded by text, or only the table by itself on the page (Fig.~\ref{fig:generated-image}~(a) and~(b)).

\subsection{Bias Mitigation}
To address synthetic data limitations, we implement:
\begin{itemize}
    \item Style randomization: 15\% probability of applying non-Western table features (right-to-left text, vertical headers)
    \item Noise injection: Scanning artifacts simulated through Poisson noise ($\lambda=0.1$) and random perspective transforms
    \item Color variation: 10\% of tables use light blue/red backgrounds instead of standard white
\end{itemize}

\subsection{Evaluation Methods}

\subsubsection{Metric: Bitwise XOR}
The primary evaluation metric is the pixel-wise \emph{XOR error rate}. Alternative region metrics such as Intersection--over--Union (IoU), F1 score, or mean Average Precision (mAP) are widespread in object detection; however, they summarize performance at the bounding-box level and are less sensitive to the thin borders that delineate table cells. Preliminary experiments showed that small border shifts can leave IoU nearly unchanged while noticeably increasing visual error. We therefore focus on XOR, which counts every misclassified pixel and is thus more discriminative for table boundaries. Let $P$ be the binary prediction mask produced by the network and $G$ the corresponding ground-truth mask; both have the same spatial resolution $H\times W$.

\begin{figure}
    \centering
	  \includegraphics[width=.8\linewidth]{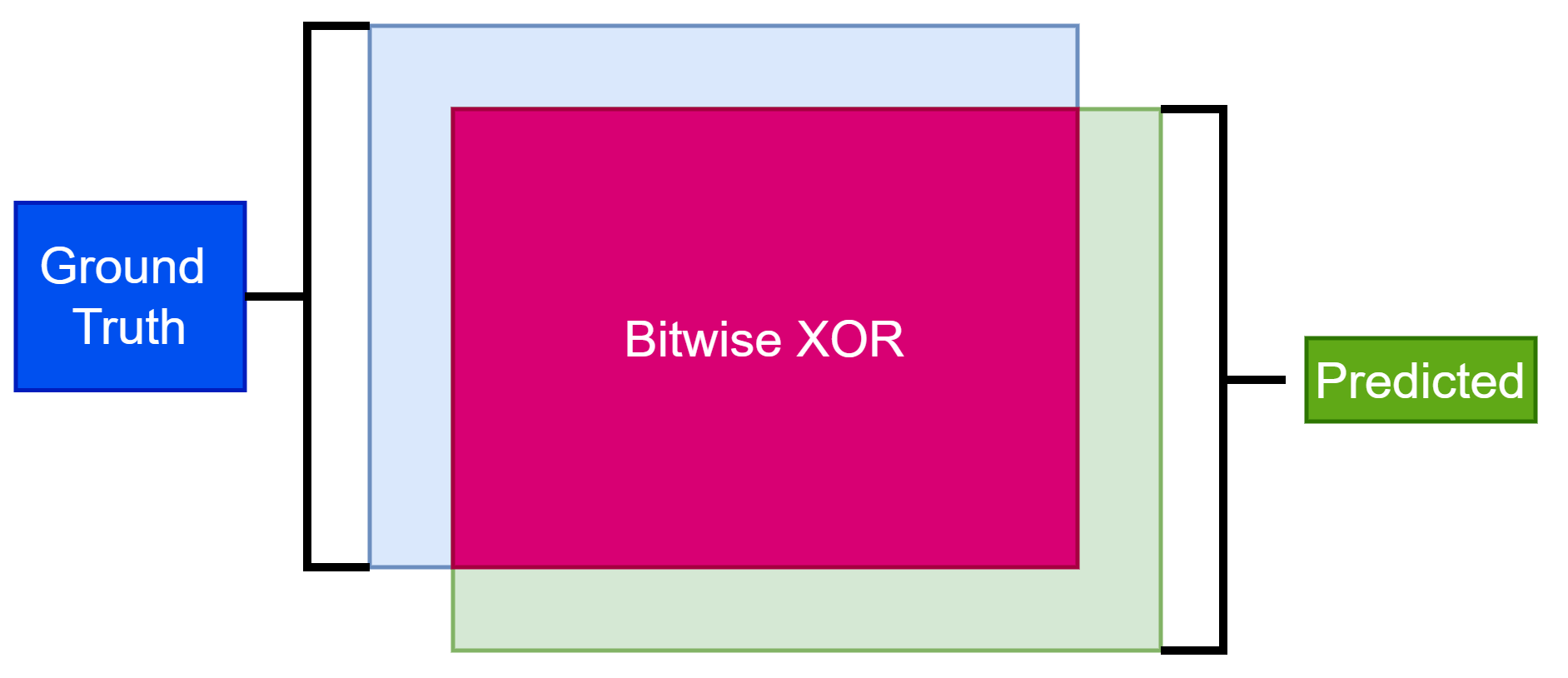}
    \caption{Example Result after Performing Bitwise XOR.}
\end{figure}

\begin{equation}
    \text{XOR Error Rate} = \frac{|P \oplus G|}{H \times W}
\end{equation}

\section{Implementation}\label{sec:implementation}

\subsection{TableNet Module}
The implementation of TableNet~\cite{paliwal2019tablenet} is used in this work. TableNet is a TensorFlow-based model~\cite{tensorflow} for table and column segmentation in grayscale images, trained on the Marmot dataset. To ensure the uniform input size of 256x256 pixels for images and 1024x1024 pixels for masks, the images are preprocessed through resizing and normalization.

TableNet's architecture utilizes a pre-trained VGG-19 base for feature extraction from RGB images of resolutions 1024x1024 or 256x256. For each corpus---both the synthetic images introduced in this work and the original Marmot pages---we randomly split the data, using 90\% of the images for training and reserving the remaining 10\% for testing. Two decoder paths, for table and column segmentation, are incorporated with convolutional and upsampling layers, along with concatenation of feature maps from VGG-19.

Training minimizes segmentation error using Sparse Categorical Crossentropy with L2 regularization ($\lambda=0.001$) to prevent overfitting. We employed gradient clipping (threshold=1.0) to stabilize training. Adam optimizer with a learning rate of 0.0001 and epsilon value of $1e^{-08}$ is used for weight adjustment. Training progress is monitored with DisplayCallback for visualization, ModelCheckpoint, and EarlyStopping for model saving and early stopping based on validation loss.

In summary, this module implements TableNet with TensorFlow, utilizing TensorFlow functions for preprocessing, VGG-19 base for feature extraction, and Adam optimizer for training. Callbacks enable monitoring of training progress for optimal segmentation results.

\subsection{Code and Data Availability}
Our complete implementation and synthetic dataset are publicly available at our code repository~\cite{github_repo}. The repository contains all data generation scripts, preprocessing utilities, trained models, and evaluation code used in this work. The dataset includes samples of all table styles discussed in this paper.

\subsection{Predict Module}
This module implements a TensorFlow prediction pipeline for table and column mask prediction on PNG images. PNG format is chosen over JPEG and BMP due to its lossless compression, preserving image quality without introducing artifacts. The pipeline utilizes a custom-trained TableNet model to segment tables and columns. By providing a sample folder path and image dimensions as inputs, users can utilize the pipeline for predicting masks and visualizing results. This end-to-end solution enables segmentation of tables and columns in image data, facilitating tasks like document processing and data extraction.

\section{Results and Discussion}\label{sec:results}

The epoch ranges were determined through progressive validation: although the 256\texttimes256 models could be trained with a larger batch size (256 instead of 64 for 1024\texttimes1024) thanks to their lower memory footprint, they still required more training epochs to converge. We attribute this to the reduced spatial detail available at lower resolutions, which slows down learning. Conversely, the higher-resolution models benefited from early stopping to prevent overfitting. This section presents an analysis of the predicted images generated by the TableNet model at two resolutions: 256\texttimes256 and 1024\texttimes1024. The goal is to evaluate the accuracy and effectiveness of these two models in table detection. Both models were trained for several sets of epochs to avoid overfitting.

\begin{figure}
    \centering
	  \includegraphics[width=.8\linewidth]{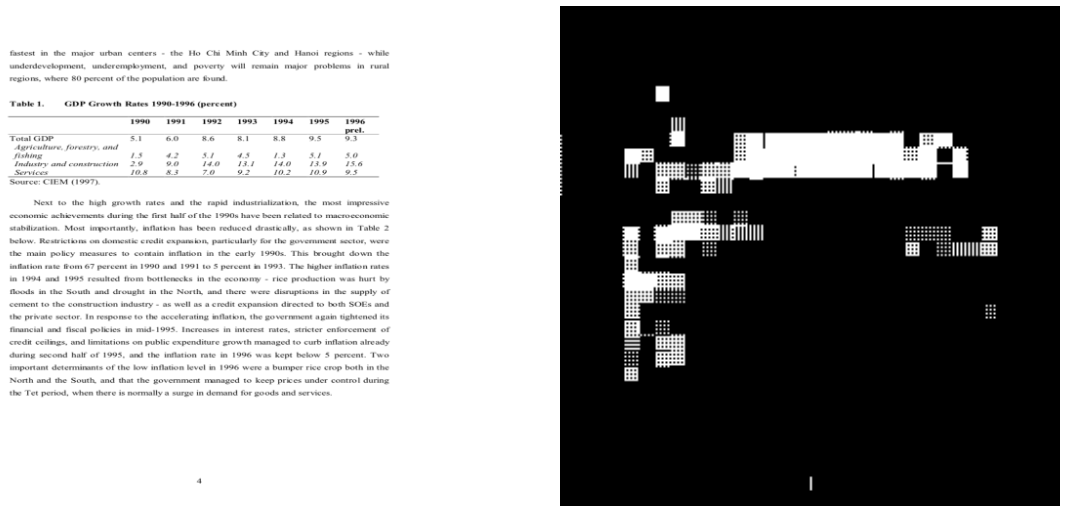}
    \caption{Sample Image from Marmot Dataset and Predicted Mask with 256\texttimes256 Model}
    \label{fig:256-marmot}
\end{figure}

The performance of the 256\texttimes256 model was evaluated at several epoch intervals. The lowest XOR error rate on the Marmot dataset was \textbf{9.18\%} at 500 epochs. On our synthetic dataset, the error rate continued to decrease with additional training, reaching \textbf{4.04\%} at 1540 epochs, suggesting the model benefits from extended training on synthetic data (Table~\ref{tab:xor-result}). A sample result is shown in Fig.~\ref{fig:256-marmot}.

\begin{figure}
    \centering
	  \includegraphics[width=.8\linewidth]{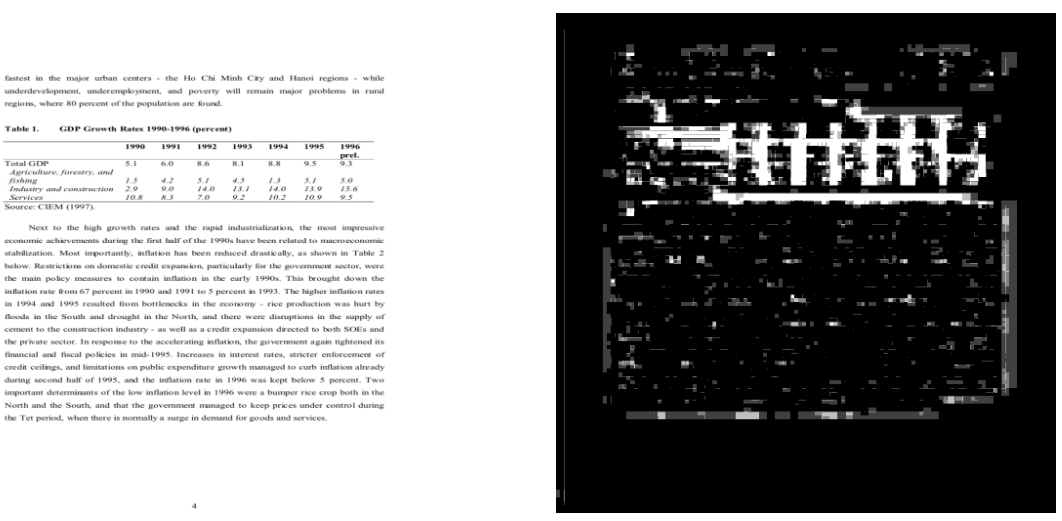}
    \caption{Sample Image from Marmot Dataset and Predicted Mask with 1024\texttimes1024 Model}
    \label{fig:1024-marmot}
\end{figure}

\begin{figure}
    \centering
	  \includegraphics[width=.8\linewidth]{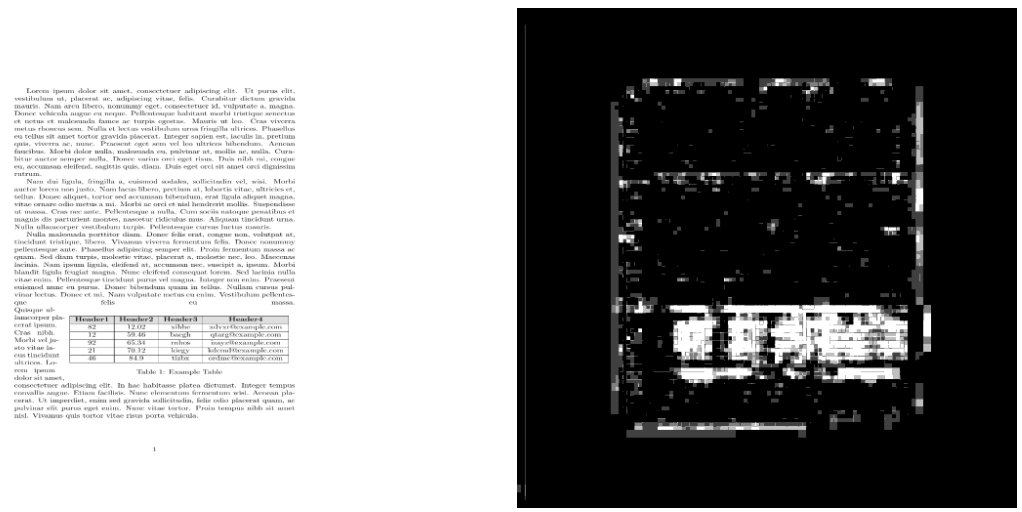}
    \caption{Sample Image from Novel Dataset and Predicted Mask with 1024\texttimes1024 Model}
    \label{fig:1024-novel}
\end{figure}

Increasing the input resolution to 1024\texttimes1024 impacted performance. The model achieved its best result on the Marmot dataset at 150 epochs with a \textbf{13.83\%} error rate, while performance degraded at higher epochs. On our synthetic dataset, the 1024\texttimes1024 model achieved its lowest error rate of \textbf{4.33\%} at 760 epochs (Table~\ref{tab:xor-result}). Sample predictions are shown in Fig.~\ref{fig:1024-marmot} and Fig.~\ref{fig:1024-novel}.


\begin{table}
    \centering
    \caption{Comparative analysis of XOR error rates (\%) across model configurations and datasets. The lowest error for each model resolution and dataset combination is shown in bold.}
    \label{tab:xor-result}
    \begin{tabular}{c|c|c}
        \hline
        Model & Marmot & Novel (Synthetic) \\
        \hline
        256\texttimes256 (240 epochs) & 12.14\% & 10.50\% \\
        256\texttimes256 (500 epochs) & \textbf{9.18\%} & 5.41\% \\
        256\texttimes256 (1540 epochs) & 11.69\% & \textbf{4.04\%} \\
        \hline
        1024\texttimes1024 (150 epochs) & \textbf{13.83\%} & 5.75\% \\
        1024\texttimes1024 (260 epochs) & 14.33\% & 4.59\% \\
        1024\texttimes1024 (760 epochs) & 13.91\% & \textbf{4.33\%} \\
        \hline
    \end{tabular}
\end{table}

The results reveal a mixed influence of input resolution. On the synthetic corpus, the lower 256\texttimes256 resolution yields a slightly better result (4.04~\%) than the 1024\texttimes1024 model (4.33~\%). In contrast, the Marmot benchmark performs better at the lower resolution, with the best 256\texttimes256 model reaching 9.18~\% versus 13.83~\% for its 1024\texttimes1024 counterpart. Hence, higher resolution does not universally improve performance on either synthetic or real-world data in our experiments.

The findings also indicate that optimal training epochs vary between datasets and resolutions.  While the synthetic dataset benefits from extended training (up to 1540 epochs for 256\texttimes256), the Marmot dataset shows signs of overfitting beyond 500 epochs for lower-resolution models.

The TableNet model, trained with a 256x256 input size, exhibits limited accuracy in predicting table regions. In some instances, it fails to detect table regions entirely. Reducing the training epochs improves detection, albeit with noisy borders and white space in the predicted masks. Conversely, the 1024x1024 model delivers crisper predictions on the synthetic corpus---reflected in its lowest 4.33~\% XOR error---yet it underperforms on the Marmot benchmark when compared with the 256x256 counterpart. Post-processing (noise removal and small-object filtering) further refines the high-resolution masks but does not overturn this overall trend. These mixed results highlight that higher input resolution alone is not a universal solution and that domain differences between synthetic and real data play a decisive role. Bitwise XOR error rate emerges as a crucial metric, quantifying pixel-wise disparities between ground truth and predicted masks, thus assessing misclassified pixels.

\section{Conclusion}\label{sec:conclusion}

This paper introduced a fully automated pipeline for generating synthetic document images, designed to address the scarcity of annotated data for table detection. Our primary contribution is a framework that not only reduces manual annotation effort but also enables a controlled and systematic evaluation of detection models. By training and testing TableNet on our generated corpus, we conducted a detailed analysis of input resolution, revealing that higher resolution (1024\texttimes1024) does not universally outperform lower resolution (256\texttimes256), particularly when transferring to real-world data like the Marmot benchmark.

This work establishes a reproducible baseline and demonstrates that a synthetic-to-synthetic evaluation approach provides crucial insights into architectural dependencies, independent of domain-specific noise. Building on this foundation, future work will extend our framework to benchmark modern architectures like TableTransformer and Cascade Mask R-CNN. We also plan to expand the data generator to produce more complex layouts, integrate OCR for end-to-end extraction, and enhance TableNet with a dedicated row identification branch to move towards full structure recognition.

Our synthetic data framework and all associated code are publicly available~\cite{github_repo}.

\bibliographystyle{IEEEtran}

\end{document}